\documentclass[twoside,twocolumn]{article}

\usepackage[scaled=.95]{newtxtext,newtxmath} 
\usepackage[T1]{fontenc} 
\usepackage[utf8]{inputenc} 
\usepackage[polish,slovene,english]{babel} 
\usepackage{microtype} 
\usepackage{graphicx} 
\usepackage{makecell} 
\usepackage{tabularx} 
\usepackage{booktabs}
\usepackage{bchart} 
\usepackage{multirow} 
\usepackage[title]{appendix}
\usepackage{tcolorbox}
\usepackage{svg}

\usepackage{cuted} 
\usepackage[hang]{footmisc} 
\usepackage[hmarginratio=1:1,margin={1.9cm,1.9cm},top=26mm,bottom=26mm,columnsep=20pt]{geometry}
\usepackage{caption} 
\usepackage{threeparttable}
\usepackage{array}
\usepackage{lettrine} 

\usepackage{enumitem} 
\setlist[itemize]{noitemsep} 

\usepackage{abstract} 

\usepackage{titlesec} 
\titleformat{\section}[block]{\large\bfseries}{\thesection}{1em}{\MakeUppercase}{} 
\titlespacing*\section{0pt}{6pt}{5pt}
\titleformat{\subsection}[block]{\large\bfseries}{\thesubsection}{1em}{} 
\titlespacing*\subsection{0pt}{6pt}{5pt}
\titleformat{\subsubsection}[runin]{\normalsize\itshape}{\thesubsubsection}{1em}{} \titlespacing*\subsubsection{0pt}{6pt}{5pt}

\usepackage{fancyhdr} 
\fancypagestyle{specialfooter}{%
  \fancyhf{}
  
  \fancyfoot[L]{©2026 Copyright held by the authors. This is the authors' version of the work. It is posted here for your personal use. Not for redistribution.}
}

\usepackage{titling} 
\usepackage{hyperref} 

\usepackage[round]{natbib} 
\newlength{\bibitemsep}
\newlength{\bibparskip}
\let\oldthebibliography\thebibliography
\renewcommand\thebibliography[1]{%
  \oldthebibliography{#1}%
  \setlength{\parskip}{\bibitemsep}%
  \setlength{\itemsep}{\bibparskip}%
}

\title{MAP: A Benchmark on Multimodal Accessibility Planning for Real World Places} 
\author{%
\textsc{Jason Armitage} \\
\normalsize University of Zurich \\
\normalsize Switzerland
\and 
\textsc{Ioannis Tsochantaridis} \\
\normalsize Google DeepMind
\and 
\textsc{Linda Mazzone} \\
\normalsize University of Zurich \\
\normalsize Switzerland
\and 
\textsc{Chuqiao Yan} \\
\normalsize University of Zurich \\
\normalsize Switzerland
\and 
\textsc{Srini Narayanan} \\
\normalsize Google DeepMind
\and 
\textsc{Sarah Ebling} \\
\normalsize University of Zurich \\
\normalsize Switzerland
}
\date{} 

\begin{document}

\maketitle
\pagestyle{empty} 
\thispagestyle{specialfooter}

\begin{abstract}
\vspace*{-.9em}
\noindent 
We introduce MAP, the first benchmark to evaluate multimodal AI systems as assistants for users with accessibility requirements when planning visits to places in the real world. In our evaluation, systems are presented with requests to verify or recommend a point of interest meeting an accessibility requirement. MAP contains two novel assessments: Claim verification for accessibility planning assesses if information on places and stated accessibility features is supported and identifies places that satisfy requested accessibility features. Visual evidence retrieval for accessibility planning checks if a multimodal AI system can select visual evidence for the requested place and accessibility feature. Our methodology supports comparison of AI systems in a setting where place information and accessibility information can change over time by evaluating systems and refreshing ground truth data at scheduled times. The benchmark is based on automatic rating and human rating for a proportion of responses.
\end{abstract}



\begin{figure*}[htbp]
    \centering
    \includegraphics[width=\textwidth, alt={This figure shows the workflow of two evaluation tasks in this MAP benchmark. One is for Claim Verification, and the other one is for Visual evidence retrieval. Under each task, there is an example of the prompt, the response and the score from evaluations.}]{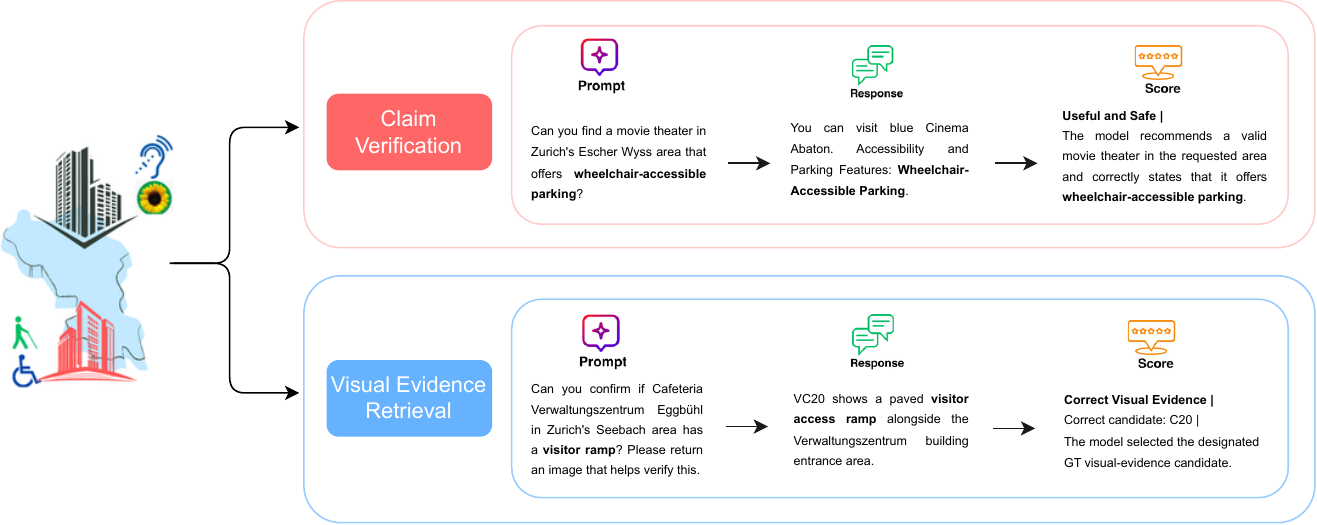}
    \caption{Overview of the two evaluation tasks in MAP: (1) Claim verification evaluates the correctness of responses on accessibility-related claims and provides a justified assessment. (2) Visual evidence retrieval evaluates whether a model can find relevant visual evidence that supports an accessibility query. The two tasks are detailed in Subsection~\ref{subsec:claim_Verification} and Subsection~\ref{subsec:visual_evidence_retrieval}.}
    \label{fig:MAP Workflow}
\end{figure*}

\section{Introduction}
\label{sec:intro}

Persons with disabilities\footnote{We use the term ``disability'' in line with the International Classification of Functioning, Disability and Health (ICF) to denote the result of interactions between an individual's impairment and restricted access to society.} often encounter barriers when planning activities and navigating urban areas, in many cases due to the complexity and fast-changing environment of urban areas \citep{ai7060206, karki2025omniacc}. They often rely on specific information to support them in planning such activities. For example, persons with mobility disabilities may look for places with larger spaces for wheelchair maneuvering or step-free entrances; persons with visual disability might require information about tactile guidance, audible signals, or lighting conditions; and persons with hearing disability may seek visual information. This accessibility information is often incomplete, distributed across different sources, or outdated. As a consequence, planning activities and navigating in the real world remain challenging for persons with disabilities.

In the recent past, multimodal AI systems -- specifically Large Language Models (LLMs) with vision input support -- have been applied as tools to assist with planning activities in the real world and have shown promising potential \citep{he2024chatgptassistvisuallyimpaired}. However, the extent to which these systems can provide \textbf{safe} and \textbf{useful} assistance remains unclear. In this work, we develop a novel evaluation framework to investigate this challenge. 

Multimodal Accessibility Planning (MAP) is a benchmark for evaluating large multimodal AI systems as assistants to users with accessibility requirements when finding places to visit in the real world. The benchmark evaluates the abilities of systems to provide useful and accurate information for places referenced in prompt responses. Assessments are performed by comparing this information to ground truth (GT) data on accessibility features. The comparison is partly based on large multimodal AI systems and checks by humans.

The evaluation addresses a real-world environment in which accessibility information and place information can change over time. Relevant changes include, for example, updates to physical access, new or closed places, and revised evidence about features such as step-free entrance, sensory provisions, or wheelchair-accessible spaces. MAP uses refreshed ground truth data and fixed scoring procedures to support comparisons between systems over time.

We identify three contributions of our research:
\begin{itemize}
\item \textbf{Multimodal Accessibility Planning (MAP) benchmark:} We introduce MAP, a benchmark for evaluating multimodal AI systems as assistants for users with accessibility requirements when planning visits to real-world places. The benchmark covers accessibility requirements associated with mobility, vision, hearing, and cognition.

\item \textbf{Two accessibility planning evaluations:} We introduce claim verification for assessing points of interest (POI) and accessibility claims in responses to \textit{feature-check} and \textit{search-by-feature} prompts as well as visual evidence retrieval for assessing whether systems can select images showing the requested POI and accessibility feature area.

\item \textbf{Ground truth for real places:} We construct a reference set of time-based data on POIs and accessibility features by integrating place and accessibility information from multiple public sources. The workflow combines automated construction by agents with human reviews to map evidence to a single accessibility schema, track provenance, and resolve source conflicts.

\end{itemize} 


\section{Background}
\label{sec:review}

To provide context for MAP, we present reviews of prior work on multimodal and factuality benchmarks (see Subsection \ref{subsec:benchrev}), AI methods proposed in the area of accessibility (see Subsection \ref{subsec:aairev}), and evaluation methods relevant to this research (see Subsection \ref{subsec:assess}).

\subsection{Benchmark Design}
\label{subsec:benchrev}

MAP evaluates multimodal AI systems, including Qwen~\citep{bai2025qwen25vltechnicalreport}, Gemma~\citep{team2026gemma}, GPT~\citep{openai2025gpt52systemcardupdate}, Claude Sonnet ~\citep{anthropic2026claudesonnet46}, Grok~\citep{xai2026grok45}, and Gemini~\citep{team2023gemini}, on their ability to provide accurate information about accessibility features of real-world locations. Previous benchmarks evaluated the performance of multimodal AI systems across a range of multimodal tasks, such as MMMU~\citep{yue2024mmmu} and MMBench~\citep{liu2024mmbench}, LiveBench~\citep{white2025livebench}, or MMIE~\citep{xia2025mmie}. Among them, factual accuracy has become one of the most important evaluation aspects. For example, FACTS~\citep{jacovi2025facts} tracks progress in factuality with a focus on long-form generation, and MFC-Bench~\citep{wang2024mfc} evaluated three fact-checking tasks.

Given the importance of factual reasoning in multimodal systems, evaluation has extended to spatial and navigation-related tasks, such as CitySeeker~\citep{wang2025cityseeker}, V-IRL~\citep{yang2024v}, or NavBench~\citep{qiao2026navbench}. These contributions provide useful insights into model performance; however, they do not reflect the complexity of the real world. Instead, they focus on static scenarios in which the place or POI has been pre-selected and the GT information remains fixed over time. Our MAP benchmark addresses this limitation by assessing whether multimodal AI systems can provide \textit{up-to-date} and factually correct accessibility information in an urban centre. 

A common paradigm used by previous benchmarks for constructing underlying datasets is to ground evaluation data in real-world information. For example, LiveBench~\citep{white2025livebench} constructs questions from recently available real-world information, while CitySeeker~\citep{wang2025cityseeker} and V-IRL~\citep{yang2024v} leverage real-world geospatial data and scene imagery. These approaches demonstrate how evaluation data can be grounded in real-world contexts. In our MAP benchmark, we similarly collect accessibility information from real-world online sources, including both textual and visual evidence, to construct the ground-truth data.

\subsection{Assessing AI Systems on Accessibility}
\label{subsec:aairev}

Evaluation of AI systems on accessibility-related tasks is a young area of research. Prior work has investigated accessibility across different domains and disability groups. For example, Accessibility SCOUT~\cite{huang2025accessibility} proposed an approach to auditing the accessible mobility features of built environments. StreetReaderAI~\cite{froehlich2025making} focused on enabling blind users to virtually navigate Google Street View through context-aware multimodal AI systems. \citet{he2025using} evaluated the capability of ChatGPT-4o to answer micro navigation queries for blind and low-vision users using scene images and corresponding human-generated descriptions. Similar evaluation studies have been conducted in the context of hearing disability. \citet{munoz2026parent} evaluated ChatGPT-4o mini and ChatGPT Pro responses on childhood hearing loss. \citet{pourhoseingholi2025validity} conducted a systematic review on the validity, reliability, and readability of four AI chatbots (ChatGPT-3.5, Bing, Gemini, and Perplexity) answering hearing-related health questions. All of these studies are constrained to one single disability type and a specific application domain, which may not be representative of the real and dynamic world. In this work, we extend beyond this limitation and provide a broader evaluation framework that contains multiple disability types and provides updated accessibility information about real-world locations relevant to diverse features.

\subsection{Evaluation Methods and Metrics}
\label{subsec:assess}

In multimodal recommendation tasks, the most commonly used metric is top-$N$ accuracy, which measures whether a ground-truth item is ranked within the top $N$ predictions, as used in AgentRecBench~\citep{shang2026agentrecbench} and AgentSelect~\citep{shi2026agentselect}. In multimodal verification tasks, accuracy and F1 score are commonly used, as, for example, in FACTS \citep{jacovi2025facts}, RW-post~\citep{xu2026rw}, or MUSCICLAIMS~\citep{lal2025musciclaims}. 

Automatic evaluation metrics are often complemented by human evaluation, as in SceneScout~\citep{jain2025scenescout}, GuideDog~\citep{kim2025guidedog}, and \cite{karamolegkou2025evaluating}. Due to the high cost of human evaluation, LLM-as-a-judge has become a widely used evaluation strategy, and previous work has shown results comparable with human evaluation \citep{laskar2025judging,gu2026survey,pu2025judge}.


\section{The MAP Benchmark}
\label{sec:benchmark}

MAP evaluates large multimodal AI systems as assistants for users with accessibility requirements when planning visits to real places. Participants in one of several workshops carried out with target users  noted that accuracy is a primary concern and that they preferred detailed explanations supported by (possibly visual) evidence. The MAP benchmark is designed specifically to address these needs.
The benchmark consists of two evaluations. \textbf{Claim verification for accessibility planning} assesses whether system responses to feature check and search-by-feature prompts make POI and accessibility claims that are supported by GT data. \textbf{Visual evidence retrieval for accessibility planning} assesses the ability of a system to select visual evidence for accessibility features of places from a candidate image set.

MAP evaluates AI system responses against GT features. Each system receives a structured prompt specifying a place or place type, a city area, and one or more accessibility features. The system's response is then checked against the GT representation of POIs and their associated accessibility features. 

\subsection{Structured Prompts}
\label{sec:prompts}

\begin{figure}[hbt!]
    \centering
    \includegraphics[width=\columnwidth, alt={This figure shows the selection process for visual evidence retrieval tasks. Three steps are illustrated: prompts, candidate sets with 20 images, and selected four key candidates.}]{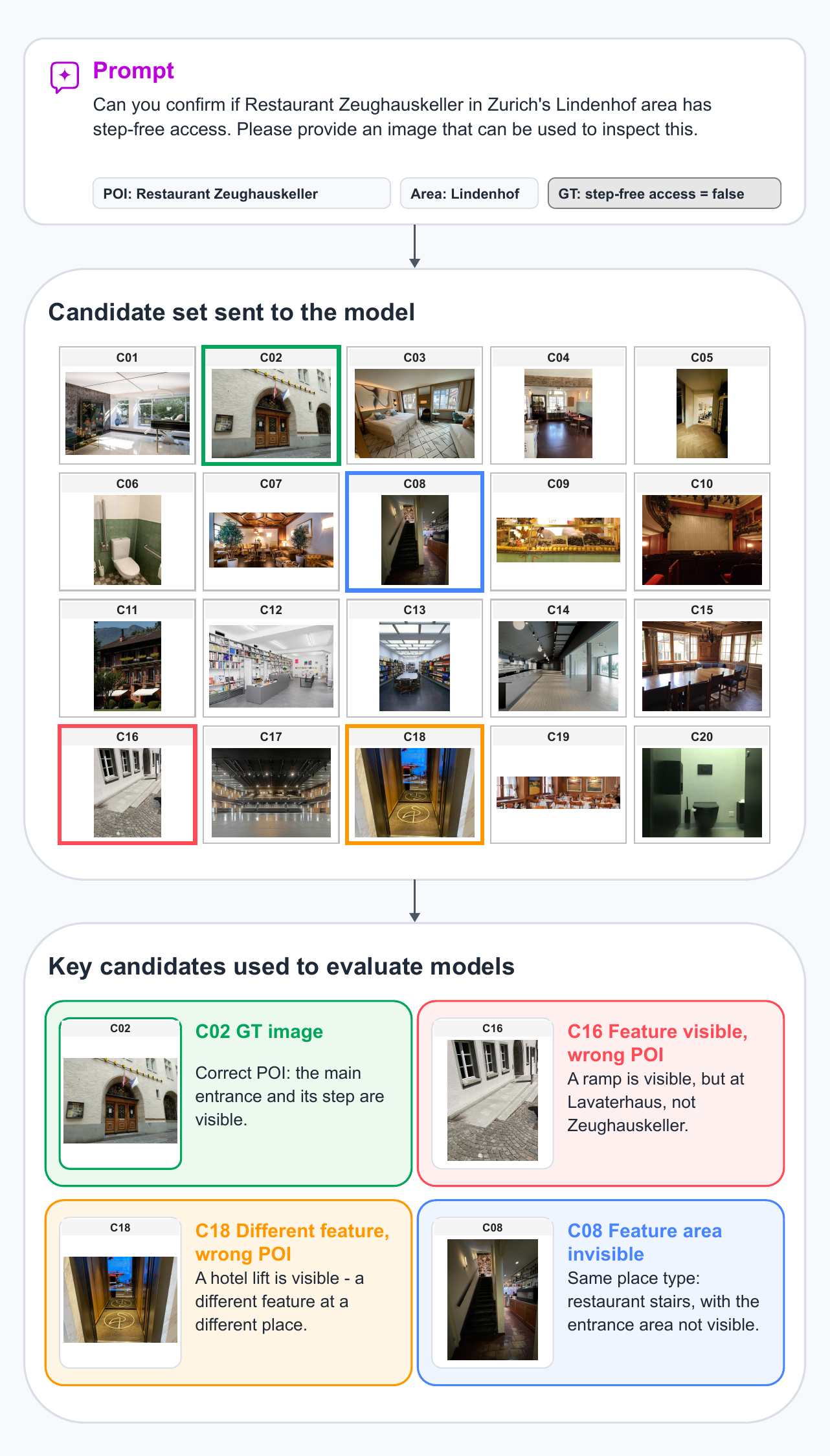}
    \caption{Inputs for the visual evidence retrieval task described in Subsection~\ref{subsec:visual_evidence_retrieval}. Each prompt is paired with a set of $n=20$ candidate images and the evaluated model must select one candidate ID or \texttt{None}. The four key image candidate types are the GT sample in green and three controlled distractors in red, gold, and blue.}
    \label{fig:ic_candidate_set}
\end{figure}

Each prompt links one or more access requirements to a named POI and a place type in a
city area. This first version of the MAP benchmark focuses on the geographical context of Zurich, and the prompts are generated based on the following dimensions:

\begin{itemize}
    \item \textbf{Disability type}: Four disability categories are included in this first version of the benchmark: mobility disability, vision disability, hearing disability, and cognitive disability.
    \item \textbf{Accessibility feature}: This is the element describing the requirement, reflecting both the disability type and the specific need. Examples include ``step-free entrances'' for mobility disabilities, ``tactile paving paths to entrances'' for visual disabilities, ``visual fire alarms and emergency alerts'' for hearing disabilities, and ``simplified instructions'' for cognitive disabilities. Prompts may also include co-occurring features.
    \item \textbf{Question type}: Two types of questions are evaluated as forms of claim verification. \textit{Feature check} questions ask whether a named POI meets those requirements. \textit{Search-by-feature} questions ask the model to suggest one or more locations that satisfy a given set of accessibility requirements. For example, ``Can you find out whether Stadthaus Zürich in the Lindenhof area of Zurich features a hearing loop?'' is a feature check question, while ``Could you find a sports activity location in the Höngg area of Zurich equipped with clear signage and a tactile guidance system?'' is a search-by-feature question. In visual evidence retrieval questions, each prompt is paired with a candidate set of 20 images containing the GT image and controlled distractors (see Figure \ref{fig:ic_candidate_set}). The evaluated system receives the candidate images in a fixed order with candidate identifiers and returns a selected candidate identifier. The assessment checks whether the image refers to the correct POI and whether the relevant area for the requested accessibility feature is visible. 
\end{itemize}

Depending on the question type, the prompt text includes either a place type or a location:

\begin{itemize}
    \item \textbf{Place type}: A category of place (e.g., museum, restaurant, park), used in search-by-feature questions where the model needs to recommend a place matching that category and the accessibility requirements.
    \item \textbf{Location}: A specific place (e.g., ``PBZ Library in Zürich Oerlikon'', ``Zürich HB''), used in feature check questions where the model checks whether that place satisfies the requirements.
\end{itemize}

The following gives one prompt example for each of the two evaluation tasks claim verification and visual evidence retrieval, with the former task consisting of two subtasks, feature check and search-by-feature (see Appendix~\ref{sec:app_2} for further examples):\\
\textbf{Claim verification:} \\
\textit{Feature check:} ``Could you confirm whether Science Pavilion UZH in Zurich's Oberstrass area offers wheelchair-accessible parking?''\\
\textit{Search-by-feature:} ``Can you find a movie theater in Zurich's Escher Wyss area that offers wheelchair-accessible parking?''\\
\textbf{Visual evidence retrieval:} ``I need to know whether Landesmuseum Zürich in Zurich's City area features clear signage. Please provide an image showing the signage.''

The prompts were produced using Gemini 3.1 Pro Preview\footnote{\url{https://gemini.google.com/app}}. Within the prompt, we assemble the above dimensions into separate subsections and instruct the model to randomly select, pick, or combine elements from each. The distribution and combination percentages for these elements were derived from analysis of our GT data. Of the $4,514$ places tracked, the proportions with evidence for at least one accessibility feature associated with each disability category are $82.7\%$ for mobility disability, $11.7\%$ for vision  disability, $6.2\%$ for cognitive disability, and $4.4\%$ for hearing disability. To limit the model from generating unrealistic prompts, the accessibility feature distribution was mapped to match the distribution observed in the GT data (see Section~\ref{sec:gtdata}).

In addition, the last subsection of the prompt specifies particular elements we want to include. First all generated prompts are limited to Zurich local areas. Second the references to accessibility features must be specific and avoid general descriptions. In the case of prompts for visual evidence retrieval, the text refers to the GT image containing the feature at the POI. The result is three prompt sets covering 59 accessibility features over 257 place types located in 34 local areas.

\subsection{Ground Truth Data}
\label{sec:gtdata}

MAP contains GT data for a finite set of POIs. Each POI is represented with \textbf{place attributes} such as name, location, place type, website, and \textbf{accessibility features}. The latter includes environmental features with a link to accessibility such as accessible entrances, hearing loops, tactile guidance, and quiet areas.

We run an automated workflow to generate a first iteration of the GT from sources that cover place information and accessibility-specific information. These include aggregated location and accessibility services -- notably Google Places and Ginto\footnote{\url{https://www.ginto.guide/}} -- together with official place websites, targeted search results, user reviews, and images of the POIs. A single reference set of data is generated by applying policies for entity matching, source selection, and conflict handling. The aim is to ensure that each POI in the GT has the most recent and relevant information available for reference. Subsequent to automated data collection and agentic assessment of information, a second iteration of the GT results from a round of reviews by humans.

\subsubsection{Places} 
The first step is to construct a single POI reference list for the target geography. The aim is to obtain complete coverage of relevant place types such as restaurants, museums, libraries, public buildings, and transport for each area in the target. In this first iteration of MAP, we focus on the city of Zurich.

The place list is built by retrieving data from multiple sources and aligning the information into POI entries. Two steps subsequent to retrieval are performed:
\begin{enumerate}
\item Identifying records that match based on name similarity, coordinate distance, address overlap, category or business-purpose agreement, and source priority.
\item Merging the information in records that are assessed as belonging to the same place and retaining as distinct records specific visitor destinations such as separate departments inside a local government building.
\end{enumerate}

\subsubsection{Accessibility Features}
MAP uses a single schema to represent accessibility information across POIs. Fields are relevant to mobility, vision, hearing, and cognition access requirements. Information from multiple sources and presented in different languages and modalities is mapped to relevant fields. For example, textual evidence for a step-free entrance or an image showing a level entrance from the street may contribute to entrance access fields.

In the final step, an agent compares sources and integrates the collected evidence using relevance, recency, source type, and agreement as criteria. The output is a single value for each field and can take on the following formats:
\begin{itemize}
\item Boolean values are used for fields such as \texttt{elevatorAvailable}.
\item Text is used for fields where multiple forms of a feature or service may exist as is the case for \texttt{signLanguageSupport}. Image descriptions and paths to the stored samples are also recorded in this form.
\item Floating-point numbers are used for fields with measurements of features such as \texttt{liftDoorWidthCm}.
\item Fields with no data and unresolved conflicts pending available information at a later date are set to \texttt{null}.
\end{itemize}

For evaluation, GT context is assembled according to the prompt type. For feature check prompts, GT context includes the specified POI, MAP accessibility fields, retained evidence, and requested feature values. For search-by-feature prompts, GT context includes the POIs eligible for the prompt, the requested place type and city area, MAP accessibility fields, and retained evidence. For visual evidence retrieval prompts, retained images are stored with image descriptions, feature visibility information, and candidate metadata recording the GT image and controlled distractor labels used for scoring.

\subsubsection{GT Construction and Updating} 
A second iteration of the GT is constructed with the assistance of humans completing sets of samples made up of an accessibility feature, an area, and -- in the case of search-by-feature prompts -- a place type. Consider a search scenario that requires information on all Swiss restaurants in a Zurich area where one candidate is missing information on the availability of wheelchair-accessible seating. A human reviewer is provided with the incomplete set to fill the missing values by inspecting official websites, search results, and images.

Records of provenance are retained with details of all sources, date and time information, scores on the criteria, and notes on logic in reconciling conflicting sources. This resource provides an audit for how the value for each field was selected. Conflicting information is also stored here to permit identifying causes of model errors.

To keep the benchmark current, GT data is refreshed shortly before scheduled evaluation runs.

\begin{table*}[hbt!]
\centering
\begin{threeparttable}
\normalsize
\setlength{\tabcolsep}{2pt}
\begin{tabular*}{\textwidth}{@{\extracolsep{\fill}}l *{8}{>{\raggedleft\arraybackslash}p{1.55cm}}}
\toprule
                                & \multicolumn{4}{c}{\textbf{No Search}} & \multicolumn{4}{c}{\textbf{Search}} \\
\cmidrule(r{0.25em}){2-5}\cmidrule(l{0.25em}){6-9}
                                & \multicolumn{1}{>{\centering\arraybackslash}p{1.55cm}}{\textbf{Safe}}
                                & \multicolumn{1}{>{\centering\arraybackslash}p{1.55cm}}{\textbf{Unverified}}
                                & \multicolumn{1}{>{\centering\arraybackslash}p{1.55cm}}{\textbf{Contradictory}}
                                & \multicolumn{1}{>{\centering\arraybackslash}p{1.55cm}}{\textbf{NU}}
                                & \multicolumn{1}{>{\centering\arraybackslash}p{1.55cm}}{\textbf{Safe}}
                                & \multicolumn{1}{>{\centering\arraybackslash}p{1.55cm}}{\textbf{Unverified}}
                                & \multicolumn{1}{>{\centering\arraybackslash}p{1.55cm}}{\textbf{Contradictory}}
                                & \multicolumn{1}{>{\centering\arraybackslash}p{1.55cm}}{\textbf{NU}} \\ 
\midrule
\textbf{Claude Sonnet 4.6}       & 1 & 898 & 1 & 0 & 69 & 742 & 85 & 4 \\
\textbf{Claude Sonnet 5}         & 0 & 900 & 0 & 0 & 85 & 730 & 85 & 0 \\
\textbf{Gemini 3.1 Pro Preview}  & 146 & 634 & 100 & 20 & 141 & 602 & 134 & 23 \\
\textbf{Gemini 3.6 Flash}        & 210 & 345 & 340 & 5 & \textbf{213} & 432 & 251 & 4 \\
\textbf{GPT-5.2}                 & 0 & 898 & 0 & 2 & 167 & 643 & 88 & 2 \\
\textbf{GPT-5.6 Terra}           & \textbf{243} & 548 & 107 & 2 & 183 & 581 & 128 & 8 \\
\textbf{Grok 4.3}                & 12 & 873 & 14 & 1 & 185 & 561 & 151 & 3 \\
\textbf{Grok 4.5}                & 166 & 638 & 94 & 2 & 189 & 497 & 211 & 3 \\
\cmidrule{1-9}
\textbf{Gemma 4 26B A4B}         & 47 & 766 & 82 & 5 & 4 & 895 & 1 & 0 \\
\textbf{Qwen2.5-VL-32B-Instruct} & 8 & 890 & 2 & 0 & 8 & 891 & 1 & 0 \\
\bottomrule
\end{tabular*}
\caption{Results on the test partition are reported for $n=900$ \textit{feature check} prompts in the claim verification task where each prompt asks the system to verify one or more accessibility features in a given place. Labels are defined in Subsection \ref{subsec:claim_Verification} and applied using auto-rating. In ``No Search'' runs, models use the original prompt only. ``Search'' permits models to retrieve online evidence. Cell entries denote response counts. \textbf{NU}: Not Useful.}
\label{table:map_claim_feature_check_test_results}
\end{threeparttable}
\end{table*}


\section{Evaluation}

In claim verification, system responses are assessed by checking whether the named POIs are present in the GT set and whether the stated features are supported by GT evidence. In visual evidence retrieval for accessibility planning, system selections are scored against the GT candidate metadata.

Closed-weight systems are evaluated through APIs and open-weight systems are assessed by loading models with the default weights from the provider. Systems are evaluated by the MAP team during a scheduled window. Existing systems are rerun following updates to the benchmark prompts and GT. Score changes and correlation with previous results aim to assess system performance over time.

\subsection{Claim Verification for Accessibility Planning}
\label{subsec:claim_Verification}
The evaluation contains two paired question types: \textit{feature check} and \textit{search-by-feature}. Feature check prompts ask the system to assess one named POI and one or more accessibility features. Search-by-feature prompts ask the system to identify places in a city area and place type that satisfy one or more access requirements. System outputs may include one or more POIs, claims about accessibility features, and source references. The evaluation checks whether each relevant POI and accessibility claim corresponds to GT evidence. Performance is reported with an output label and summary metrics calculated from the assigned labels.

\begin{table*}[hbt!]
\centering
\begin{threeparttable}
\normalsize
\setlength{\tabcolsep}{2pt}
\begin{tabular*}{\textwidth}{@{\extracolsep{\fill}}l *{12}{>{\raggedleft\arraybackslash}p{0.62cm}}}
\toprule
                                 & \multicolumn{6}{c}{\textbf{No Search}} & \multicolumn{6}{c}{\textbf{Search}} \\
\cmidrule(r{0.25em}){2-7}\cmidrule(l{0.25em}){8-13}
                                 & \multicolumn{1}{>{\centering\arraybackslash}p{0.62cm}}{\textbf{US}}
                                 & \multicolumn{1}{>{\centering\arraybackslash}p{0.62cm}}{\textbf{SU}}
                                 & \multicolumn{1}{>{\centering\arraybackslash}p{0.62cm}}{\textbf{UU}}
                                 & \multicolumn{1}{>{\centering\arraybackslash}p{0.62cm}}{\textbf{PUU}}
                                 & \multicolumn{1}{>{\centering\arraybackslash}p{0.62cm}}{\textbf{UPU}}
                                 & \multicolumn{1}{>{\centering\arraybackslash}p{0.62cm}}{\textbf{NU}}
                                 & \multicolumn{1}{>{\centering\arraybackslash}p{0.62cm}}{\textbf{US}}
                                 & \multicolumn{1}{>{\centering\arraybackslash}p{0.62cm}}{\textbf{SU}}
                                 & \multicolumn{1}{>{\centering\arraybackslash}p{0.62cm}}{\textbf{UU}}
                                 & \multicolumn{1}{>{\centering\arraybackslash}p{0.62cm}}{\textbf{PUU}}
                                 & \multicolumn{1}{>{\centering\arraybackslash}p{0.62cm}}{\textbf{UPU}}
                                 & \multicolumn{1}{>{\centering\arraybackslash}p{0.62cm}}{\textbf{NU}} \\ 
\midrule
\textbf{Claude Sonnet 4.6}       & 15 & 28 & 4 & 40 & 0 & 513 & 99 & 19 & 52 & 117 & 19 & 294 \\
\textbf{Claude Sonnet 5}         & 15 & 42 & 22 & 126 & 1 & 394 & 73 & 31 & 92 & 105 & 22 & 277 \\
\textbf{Gemini 3.1 Pro Preview}  & 93 & 5 & 57 & 212 & 40 & 193 & 89 & 6 & 64 & 87 & 44 & 310 \\
\textbf{Gemini 3.6 Flash}        & 61 & 0 & 59 & 174 & 47 & 259 & 91 & 0 & 77 & 77 & 55 & 300 \\
\textbf{GPT-5.2}                 & 17 & 76 & 14 & 85 & 0 & 408 & \textbf{141} & 2 & 35 & 126 & 30 & 266 \\
\textbf{GPT-5.6 Terra}           & \textbf{102} & 4 & 10 & 401 & 14 & 69 & 132 & 7 & 61 & 93 & 42 & 265 \\
\textbf{Grok 4.3}                & 20 & 7 & 0 & 143 & 1 & 429 & 121 & 11 & 59 & 227 & 17 & 165 \\
\textbf{Grok 4.5}                & 60 & 7 & 10 & 278 & 6 & 239 & 116 & 3 & 119 & 112 & 47 & 203 \\
\cmidrule{1-13}
\textbf{Gemma 4 26B A4B}         & 26 & 0 & 23 & 229 & 7 & 315 & 20 & 55 & 16 & 128 & 1 & 380 \\
\textbf{Qwen2.5-VL-32B-Instruct} & 8 & 6 & 4 & 123 & 3 & 456 & 8 & 7 & 3 & 31 & 0 & 551 \\
\bottomrule
\end{tabular*}
\caption{Results on the test set of the \textit{search-by-feature} prompts in the claim verification task. Requests to systems are based on identifying places of a specified type in a city area that meet specific access requirements. Performance is evaluated based on verified/supported decisions. ``No Search'' and ``Search'' depend on the access of the system to an online search component (e.g., Google Search) for retrieving evidence. Reported values are for labels applied by an auto-rater assessing responses. Output labels are denoted as follows: \textbf{US} (Useful and Safe), \textbf{SU} (Safe but Uninformative), \textbf{UU} (Useful but Unverified), \textbf{PUU} (Potentially Useful but Unverified), \textbf{UPU} (Useful but Potentially Unsafe), and \textbf{NU} (Not useful).}
\label{table:map_claim_verification_search_test_results}
\end{threeparttable}
\end{table*}

\begin{figure*}[t]
    \centering
    \includegraphics[width=\textwidth, alt={This figure contains 12 radar charts, with each showing the evaluation results from closed-weight models. More interpretation results are explained in section 4.1.1.}]{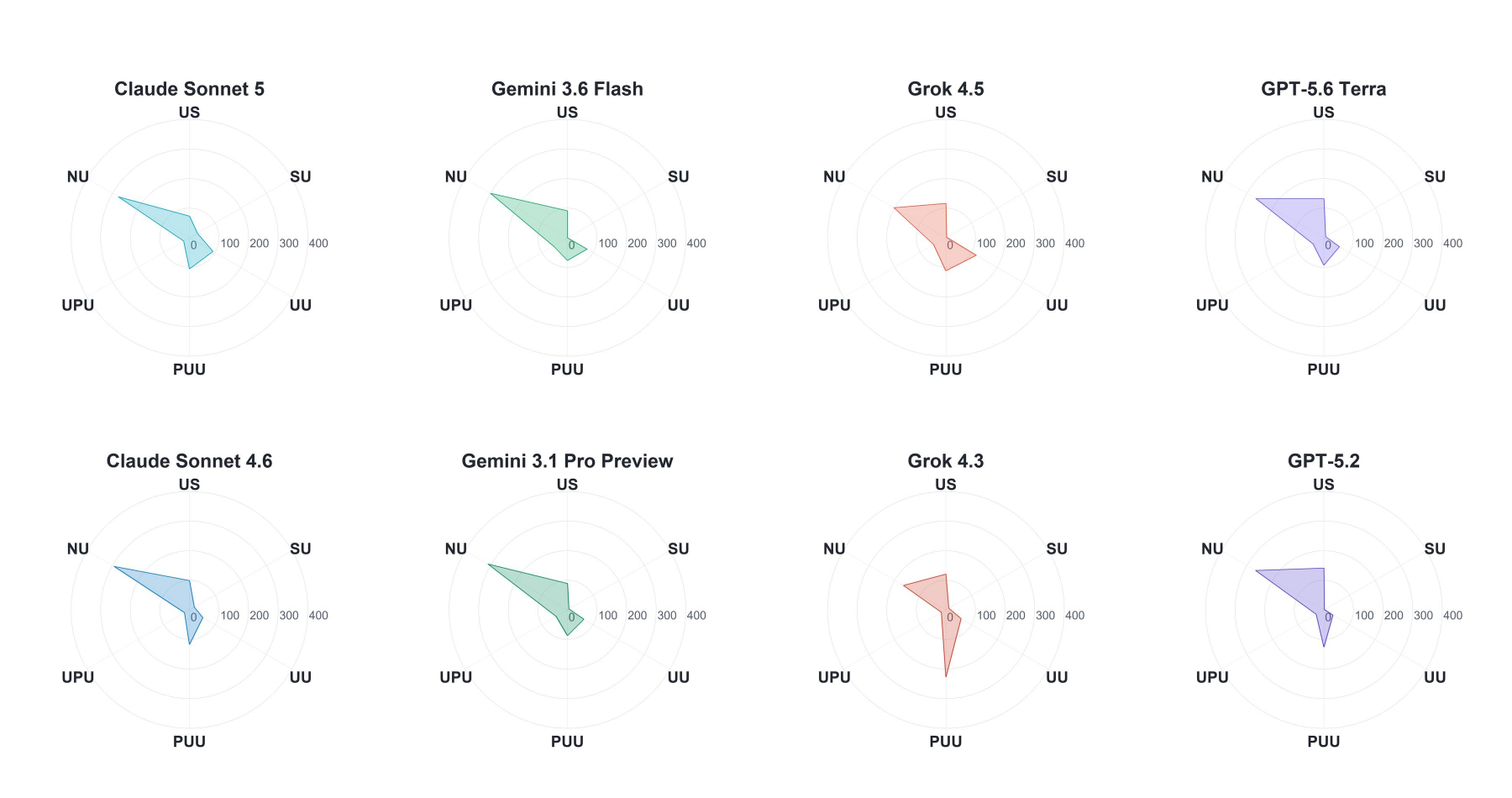}
    \caption{Results for closed-weight models  with ``Search'' enabled on the test partition of the \textit{search-by-feature} prompts indicating that US (Useful and Safe) accounts for less than a quarter of labels ($n=600$). A marginal fraction of responses by open-weights models are scored as US. Counts are provided in Table~\ref{table:map_claim_verification_search_test_results}. Output labels are denoted as follows: \textbf{US} (Useful and Safe), \textbf{SU} (Safe but Uninformative), \textbf{UU} (Useful but Unverified), \textbf{PUU} (Potentially Useful but Unverified), \textbf{UPU} (Useful but Potentially Unsafe), and \textbf{NU} (Not Useful).}
    \label{figure:map_claim_verification_search_feature_results}
\end{figure*}

Automatic rating (auto-rating) is used for full evaluation runs and human rating supports external checking of a proportion of responses. 

\subsubsection{Auto-rating}
An auto-rater performs checks against rules defined by a policy and refers to a single set of consolidated GT information to support reproducible comparisons over evaluated systems. The evaluation uses label-based scores to describe whether the system output is useful for the accessibility planning request and whether the stated accessibility information is safe to rely on. For feature check prompts, the possible labels are the following:

\begin{itemize}
\item \textbf{Safe}: The response addresses the place specified in the prompt, gives a definite answer for every requested accessibility feature, and all accessibility claims in the response are covered by and match the GT.
\item \textbf{Unverified}: The response addresses the place specified in the prompt and contains no definite accessibility claim that contradicts the GT but does not meet the requirements for Safe: Requested information is unresolved, omitted, answered only with broader accessibility information that does not include the specific feature, or is unresolved between competing sources. This label is also applied when a definite additional claim is not covered by the GT.
\item \textbf{Contradictory}: The response addresses or purports to address the place specified in the prompt, and at least one accessibility claim in the response contradicts the GT.
\item \textbf{Not Useful (NU)}: The response is absent, the model refuses to engage with the request, or the response does not address the place specified in the prompt.
\end{itemize}

For search-by-feature prompts, the output labels are defined as follows:
\begin{itemize}
\item \textbf{Useful and Safe (US)}: All recommended POIs are present in the GT, and all relevant accessibility claims in the response are supported by the GT.
\item \textbf{Safe but Uninformative (SU)}: All recommended POIs are present in the GT, but the response provides no information on accessibility features relevant to the query.
\item \textbf{Useful but Unverified (UU)}: At least one recommended POI is present in the GT and the response contains supported or unverified accessibility information, with no contradicted accessibility claim. This label applies when the response also includes a POI outside the GT or an accessibility claim that the GT can neither confirm nor contradict.
\item \textbf{Potentially Useful but Unverified (PUU)}: The response recommends one or more POIs, but none can be matched to the GT. The places are therefore unknown to the dataset, and their accessibility features cannot be verified.
\item \textbf{Useful but Potentially Unsafe (UPU)}: At least one recommended POI is present in the GT and the response contains some supported or unverified accessibility information, but also contains at least one accessibility claim that contradicts known GT information.
\item \textbf{Not Useful (NU)}: The response recommends no POI, does not address the prompt, or recommends GT POIs only with accessibility claims that contradict the GT and provide no safe accessibility utility.
\end{itemize}

The results in Table \ref{table:map_claim_feature_check_test_results} indicate that feature check prompts provide a signal on whether systems are able to turn available evidence into definite accessibility claims for a specified place. Across scored responses, ``Unverified'' is the dominant label. A tendency towards cautious or unresolved responses is consistent with vendor documentation detailing uncertainty and refusal as considerations when aligning model responses~\citep{openai2025modelspec,anthropic2026claudesonnet46}. Training and safety measures may contribute to avoiding definite claims on accessibility features in relation to evidence. Including ``Search'' changes the response profile for several closed-weight systems: Grok 4.3, OpenAI GPT-5.2, and Claude Sonnet 4.6 show higher ``Safe'' counts. ``Search'' also increases ``Contradictory'' labels for most closed-weight systems. Open-weight systems remain mostly ``Unverified'' in both conditions with retrieved evidence resulting in cautious responses more often than ``Safe'' ones. 

\begin{table}[hbt!]
\begin{center}
\begin{threeparttable}
\footnotesize
\resizebox{\columnwidth}{!}{%
\begin{tabular}{l >{\centering\arraybackslash}p{0.95cm} >{\centering\arraybackslash}p{0.95cm}}
\hline
                                & \textbf{Safe} & \textbf{Other} \\ \hline
\textbf{Claude Sonnet 4.6}       & 4 / 3 & 26 / 27 \\
\textbf{Claude Sonnet 5}         & 6 / 6 & 24 / 24 \\
\textbf{Gemini 3.1 Pro Preview}  & 9 / 10 & 21 / 20 \\
\textbf{Gemini 3.6 Flash}        & 13 / 12 & 17 / 18 \\
\textbf{GPT-5.2}                 & 5 / 6 & 25 / 24 \\
\textbf{GPT-5.6 Terra}           & 5 / 8 & 25 / 22 \\
\textbf{Grok 4.3}                & 6 / 14 & 24 / 16 \\
\textbf{Grok 4.5}                & 6 / 11 & 24 / 19 \\
\cmidrule{1-3}
\textbf{Gemma 4 26B A4B}         & 1 / 1 & 29 / 29 \\
\textbf{Qwen2.5-VL-32B-Instruct} & 4 / 4 & 26 / 26 \\ \hline
\end{tabular}%
}
\caption{Human/auto-rater counts of labels allocated to a set of responses by models with search for $n = 30$ feature check samples in the claim verification task. Results are for the validation partition and are provided for ``Safe'' and ``Other'', where ``Other'' combines ``Unverified'', ``Contradictory'', and ``Not Useful'' counts.}
\label{table:map_human_auto_binary_safe_other_results}
\end{threeparttable}
\end{center}
\end{table}

Search-by-feature prompts test a different failure mode to the above as systems have to both identify a suitable POI and make accessibility claims about the place. Results in Table \ref{table:map_claim_verification_search_test_results} indicate that access to search results increases the number of ``Useful and Safe'' responses for closed-weight systems. Over $200$ of the responses by Gemini 3.1 Pro Preview with search enabled are scored as ``Useful and Safe''. GPT-5.2 shows the largest gain in outputs labeled as ``Useful and Safe'' rising from 17 to 141. Search capability also increases the number of ``Useful but Potentially Unsafe'' outputs and suggests that systems tend to use retrieved evidence to produce more specific and decisive responses. Notwithstanding this, the prevalence of cautious responses is evident in the high number of responses labeled as ``Not Useful'' as visualised in Figure \ref{figure:map_claim_verification_search_feature_results}. 

\subsubsection{Human Assessment}
We present a comparison of human and auto-rater scoring in the claim verification task using $n = 30$ feature check prompts in Table \ref{table:map_human_auto_binary_safe_other_results}. Assessments performed by two humans are in alignment with auto-rating for GPT-5.2, Claude Sonnet 4.6, Claude Sonnet 5, Gemini 3.6 Flash, and the open-weights models. The widest gaps in label allocation are for the responses returned by Grok 4.3 and Grok 4.5. Challenges in evaluating claims in this task result from multiple features of responses including the provision of features that are not referenced in the prompt, contradictory information, and a lack of specificity (e.g., responding that a place is wheelchair-accessible to a question on whether the main entrance is wheelchair-accessible). See the appendices for details of specific challenges in scoring responses.

\subsection{Visual Evidence Retrieval for Accessibility Planning}
\label{subsec:visual_evidence_retrieval}
We introduce visual evidence retrieval for accessibility planning to assess whether multimodal AI systems can select visual evidence for accessibility features of real places. Each prompt specifies a POI, an accessibility feature, and a visitor need. The system receives a fixed set of candidate images and is asked to select the candidate containing visual evidence for the requested feature, or to select \texttt{None}. During scoring, the selected candidate is compared with the GT candidate metadata for the same POI and feature. The evaluation measures whether the image refers to the correct POI and whether the relevant physical area for the accessibility feature is visible. 

The evaluation uses a label-based score to describe whether the selected image provides visual evidence for the requested accessibility feature. The scoring compares the selected candidate identifier with the GT candidate metadata.

\begin{itemize}
\item \textbf{Correct Visual Evidence (CVE)}: The selected image shows the requested POI and the requested accessibility feature area.
\item \textbf{Incorrect Place but Correct Feature Area (IP-CFA)}: The selected image shows the requested accessibility feature area, but for the wrong POI.
\item \textbf{Incorrect Place and Feature Area (IP-IFA)}: The selected image does not provide the requested POI-feature match.
\item \textbf{No Evidence Detected (NED)}: The model selects \texttt{None}.
\end{itemize}

\begin{table}[hbt!]
\centering
\begin{threeparttable}
\footnotesize
\setlength{\tabcolsep}{1.5pt}
\resizebox{\columnwidth}{!}{%
\begin{tabular}{@{}l
>{\raggedleft\arraybackslash}p{0.9cm}
>{\raggedleft\arraybackslash}p{0.9cm}
>{\raggedleft\arraybackslash}p{0.9cm}
>{\raggedleft\arraybackslash}p{0.9cm}@{}}
\toprule
                                & \multicolumn{1}{>{\centering\arraybackslash}p{0.9cm}}{\textbf{CVE}}
                                & \multicolumn{1}{>{\centering\arraybackslash}p{0.9cm}}{\shortstack{\textbf{IP-}\\\textbf{CFA}}}
                                & \multicolumn{1}{>{\centering\arraybackslash}p{0.9cm}}{\shortstack{\textbf{IP-}\\\textbf{IFA}}}
                                & \multicolumn{1}{>{\centering\arraybackslash}p{0.9cm}}{\textbf{NED}} \\ 
\midrule
\textbf{Claude Sonnet 4.6}       & 213 & 25 & 7 & 55 \\
\textbf{Claude Sonnet 5}         & 130 & 12 & 3 & 155 \\
\textbf{Gemini 3.1 Pro Preview}  & 236 & 13 & 5 & 46 \\
\textbf{Gemini 3.6 Flash}        & \textbf{250} & 20 & 2 & 28 \\
\textbf{GPT-5.2}                 & 95 & 7 & 0 & 198 \\
\textbf{GPT-5.6 Terra}           & 228 & 34 & 10 & 28 \\
\textbf{Grok 4.3}                & 90 & 7 & 2 & 201 \\
\textbf{Grok 4.5}                & 138 & 13 & 2 & 147 \\
\cmidrule{1-5}
\textbf{Gemma 4 26B A4B}         & 105 & 37 & 6 & 152 \\
\textbf{Qwen2.5-VL-32B-Instruct} & 158 & 67 & 63 & 12 \\
\bottomrule
\end{tabular}%
}
\caption{Results on the test partition of the evaluation on the visual evidence retrieval task. Results are reported over $n=300$ test prompts and all models are run without access to search. Column headings are defined as follows: \textbf{CVE} = Correct Visual Evidence, \textbf{IP-CFA} = Incorrect Place but Correct Feature Area, \textbf{IP-IFA} = Incorrect Place and Feature Area, and \textbf{NED} = No Evidence Detected. Values are the count of scoring labels allocated by an auto-rater to the responses.}
\label{table:map_visual_test_results}
\end{threeparttable}
\end{table}

\begin{figure*}[t]
    \centering
    \includegraphics[width=\textwidth, alt={This figure contains 12 gauge charts, with each indicating the correct visual evidence results for each closed-weight model. More specific results are explained in section 4.2.}]{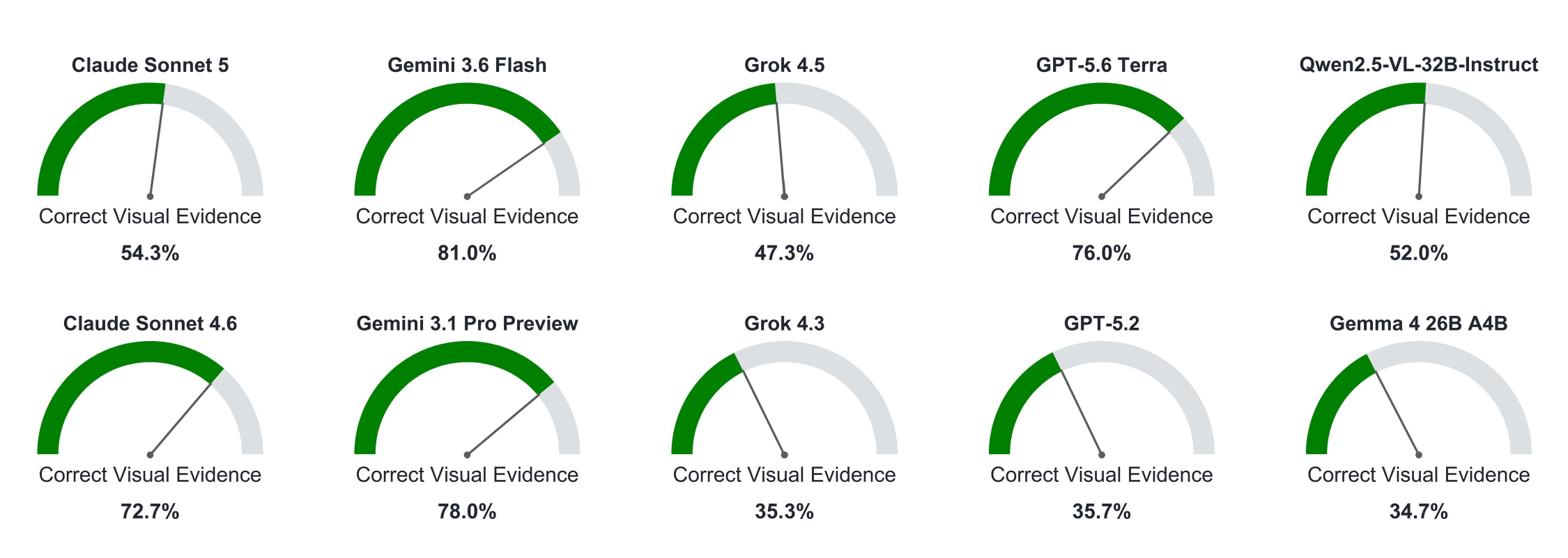}
    \caption{Correct visual evidence rates for systems on the test partition of the visual evidence retrieval task. Results indicate a model's ability to retrieve an image matching the precise accessibility feature in the requested POI. Full counts for correct and other visual evidence labels are reported in Table~\ref{table:map_visual_test_results}.}
    \label{figure:map_visual_cve_test_results}
\end{figure*}

A numeric score in $[0,1]$ is also assigned as a measure of how visible the relevant feature area is and whether it references the correct place. A low score indicates that the relevant area for the accessibility feature is not visible, the POI is wrong, or no image is retrieved. A high score indicates that the POI is correct and the relevant area is the focus of the image. For negative cases, the score still measures visibility of the relevant area. An image showing the entrance to the correct POI where steps are the focus of the image will receive a high score for a prompt asking about step-free access. Conversely, an image of the correct POI that shows only a dining area and not the entrance receives a low score for the same prompt.

The results shown in Figure~\ref{figure:map_visual_cve_test_results} and Table~\ref{table:map_visual_test_results} underline the strongest performance from Gemini 3.6, Gemini 3.1 Pro Preview, GPT-5.6 Terra, and Claude Sonnet 4.6, with each selecting Correct Visual Evidence for more than two thirds of the cases. The results are informed by two common patterns in responses: systems such as Claude Sonnet 5 tend to select \texttt{None} while other systems select controlled distractors that show the correct requested feature area for a POI that is different to the one in the prompt. Qwen2.5-VL-32B-Instruct follows the second pattern, where more selections are assigned to the two incorrect visual evidence categories. Performance also varies by feature, with selections on images of entrance and step-free access prompts producing strong results.


\section{Limitations}
\label{sec:limitations}

This work has several limitations related to the GT data, the reliance on digital sources, and the scope of the evaluation. First the GT data is formed from publicly available and digital sources. This supports scalable collection and timely updates but it also leads to uneven coverage of accessibility features. Visible features -- such as wheelchair-accessible seating, ramps, step-free entrances, and accessible parking -- appear more frequently in digital sources and formats (e.g., websites, images) than features related to, for example, hearing or cognitive disabilities. This bias is consistent with prior work reporting that accessibility assessment and documentation often emphasize physical and mobility-related features over other access requirements
\citep{carlsson2022scoping, chidiac2024accessibility}.

Digital sources may also lag behind updates in the physical world. Accessibility features can change after renovations, temporary works, venue closures, or changes in visitor support. Websites, reviews, images, and open data may not be updated at the same time. Although MAP mitigates this issue by refreshing GT data before scheduled evaluation runs, the benchmark remains dependent on the timeliness and coverage of available source evidence. 

The prompt sets also reflect decisions taken ahead of constructing this first iteration of MAP. Prompts are generated in American English and may not fully reflect language use in other regional, cultural, or multilingual contexts. Some prompts will retain synthetic wording rather than the shorter, underspecified, or error-prone requests that users may enter into systems in practice~\citep{zhao2024wildchat, liu2025creme}. 

Prompts are also bound by the set of features specified in the MAP schema and the accessibility features that can be assessed and scored. Real requests may contain features outside this coverage, temporary constraints, or include journey-level access requirements (i.e., requirements related to the route from a source to a destination). 

Finally, both the prompts and the GT data are limited to Zurich in this first iteration. Results should be interpreted as evidence for this city and not as a direct measure of accessibility planning across regions.


\section{Conclusion} 
Our evaluation assesses multimodal AI systems as assistants for users with accessibility requirements when planning visits to real places. Two novel tasks are proposed that evaluate systems on the factual accuracy of the information in responses and the ability to present visual evidence that is relevant to the user request. We enable evaluation in an open world setting and allow for dynamic changes by updating ground truth data and running assessments within specified windows.

\section{Acknowledgments}

This research was supported by a grant from Google. The authors would like to thank the team at Ginto for the insights and guidance provided during the course of this research.


{\footnotesize
\bibliography{main.bib}}

\clearpage
\appendix
\begin{appendices}

\setcounter{page}{1}
\setcounter{section}{0}
\renewcommand\thesection{\arabic{section}}

\twocolumn[{
\begin{center}
{\huge\bfseries Appendices\par}
\end{center}
\vspace{0.25in}
}]


\section{Question Types and System Prompts}
\label{sec:app_1}

Samples of prompts for each question type are presented in Table \ref{tab:map_prompt_examples}, which provides lists of elements that constitute the task prompts. At inference, these prompts are served as user requests to align with a process where human users query systems. A system prompt is provided as a separate input to instruct the model on the requirements for responding with an output suitable for evaluation. System prompts are concise and standard over evaluated models. 

\begin{strip}
\centering

\begin{tabular}{p{2.2cm} p{8.3cm} p{5.5cm}}
\toprule
\textbf{Question Type} & \textbf{Core Elements} & \textbf{Sample Prompt} \\
\midrule
Feature check &
\raggedright
\textbullet\ \textbf{Disability type}: Mobility \\ 
\textbullet\ \textbf{Accessibility feature}: Entrance step count \& Tactile paving \\ 
\textbullet\ \textbf{Place type}: Sports center \\ 
\textbullet\ \textbf{Local area}: Wollishofen \\
&
I need to check whether Strandbad Wollishofen in Zurich's Wollishofen area has information on main entrance step count and tactile paving. 
\\
\midrule
Search-by-feature &
\raggedright
\textbullet\ \textbf{Disability type}: Mobility \\
\textbullet\ \textbf{Accessibility feature}: Wheelchair accessible parking \\
\textbullet\ \textbf{Place type}: Sports center \\
\textbullet\ \textbf{Local area}: Oerlikon \\
&
Can you recommend a sports complex in Zurich's Oerlikon area with clear signage and wheelchair accessible parking? 
\\
\midrule
Visual evidence retrieval &
\raggedright
\textbullet\ \textbf{Disability type}: Blind and low vision  \\
\textbullet\ \textbf{Accessibility feature}: Contrast markings \\
\textbullet\ \textbf{Place type}: Bar \\
\textbullet\ \textbf{Local area}: Escher Wyss \\
&
Can you check if sphères in the Escher Wyss area of Zurich has contrast markings on the indoor staircase? Please provide an image for visual evidence.
\\
\bottomrule
\end{tabular}

\captionof{table}{Samples of prompts from each question type in the claim verification and visual evidence retrieval tasks.}
\label{tab:map_prompt_examples}

\vspace{1.25em}

\begin{tabularx}{\textwidth}{p{2.2cm} X}
\toprule
\textbf{Question Type} & \textbf{System Prompt} \\
\midrule
Feature check &
Answer the request as a standalone, single-turn query. Use the information sources available in this run when they can improve accuracy. Ground place information in specific, checkable information. State accessibility features when they are directly supported by available evidence. Do not use saved memory or any earlier conversation. Return text only.
\\
\midrule
Search-by-feature &
Answer the request as a standalone, single-turn query. Use the information sources available in this run when they can improve accuracy. Ground place information in specific, checkable information. State accessibility features when they are directly supported by available evidence. Do not use saved memory or any earlier conversation. Return text only.
\\
\midrule
Visual evidence retrieval &
Select the requested image from the candidate set.
Select a candidate only when it refers to the requested place and the relevant physical area is visible.
If no candidate provides adequate evidence, select NONE.
\\
\bottomrule
\end{tabularx}

\captionof{table}{System prompts used for runs listed by question type.}
\label{tab:map_system_prompts}

\end{strip}


\section{Additional Results}
\label{sec:app_2}

Results of runs on the validation partitions for the three question types in the claim verification and visual evidence retrieval tasks are presented below in Tables \ref{table:map_claim_feature_check_validation_results}, \ref{table:map_claim_verification_search_validation_results}, and \ref{table:map_visual_validation_results}. 

\begin{table*}[hbt!]
\centering
\begin{threeparttable}
\normalsize
\setlength{\tabcolsep}{2pt}
\begin{tabular*}{\textwidth}{@{\extracolsep{\fill}}l *{8}{>{\raggedleft\arraybackslash}p{1.55cm}}}
\toprule
                                & \multicolumn{4}{c}{\textbf{No Search}} & \multicolumn{4}{c}{\textbf{Search}} \\
\cmidrule(r{0.25em}){2-5}\cmidrule(l{0.25em}){6-9}
                                & \multicolumn{1}{>{\centering\arraybackslash}p{1.55cm}}{\textbf{Safe}}
                                & \multicolumn{1}{>{\centering\arraybackslash}p{1.55cm}}{\textbf{Unverified}}
                                & \multicolumn{1}{>{\centering\arraybackslash}p{1.55cm}}{\textbf{Contradictory}}
                                & \multicolumn{1}{>{\centering\arraybackslash}p{1.55cm}}{\textbf{NU}}
                                & \multicolumn{1}{>{\centering\arraybackslash}p{1.55cm}}{\textbf{Safe}}
                                & \multicolumn{1}{>{\centering\arraybackslash}p{1.55cm}}{\textbf{Unverified}}
                                & \multicolumn{1}{>{\centering\arraybackslash}p{1.55cm}}{\textbf{Contradictory}}
                                & \multicolumn{1}{>{\centering\arraybackslash}p{1.55cm}}{\textbf{NU}} \\ 
\midrule
\textbf{Claude Sonnet 4.6}       & 0 & 300 & 0 & 0 & 21 & 238 & 40 & 1 \\
\textbf{Claude Sonnet 5}         & 0 & 300 & 0 & 0 & 27 & 236 & 35 & 2 \\
\textbf{Gemini 3.1 Pro Preview}  & 56 & 212 & 28 & 4 & 57 & 185 & 56 & 2 \\
\textbf{Gemini 3.6 Flash}        & 67 & 116 & 115 & 2 & \textbf{77} & 131 & 91 & 1 \\
\textbf{GPT-5.2}                 & 0 & 300 & 0 & 0 & 46 & 221 & 33 & 0 \\
\textbf{GPT-5.6 Terra}           & \textbf{74} & 186 & 38 & 2 & 46 & 206 & 44 & 4 \\
\textbf{Grok 4.3}                & 5 & 293 & 2 & 0 & 72 & 166 & 59 & 3 \\
\textbf{Grok 4.5}                & 63 & 205 & 31 & 1 & 61 & 160 & 77 & 2 \\
\cmidrule{1-9}
\textbf{Gemma 4 26B A4B}         & 15 & 251 & 29 & 5 & 7 & 288 & 5 & 0 \\
\textbf{Qwen2.5-VL-32B-Instruct} & 1 & 291 & 0 & 8 & 14 & 278 & 6 & 2 \\
\bottomrule
\end{tabular*}
\caption{Validation results on the \textit{feature check} questions ($n=300$) in the claim verification task. Cell entries are response counts for labels allocated by the auto-rater.}
\label{table:map_claim_feature_check_validation_results}
\end{threeparttable}
\end{table*}

\begin{table*}[hbt!]
\centering
\begin{threeparttable}
\normalsize
\setlength{\tabcolsep}{2pt}
\begin{tabular*}{\textwidth}{@{\extracolsep{\fill}}l *{12}{>{\raggedleft\arraybackslash}p{0.62cm}}}
\toprule
                                 & \multicolumn{6}{c}{\textbf{No Search}} & \multicolumn{6}{c}{\textbf{Search}} \\
\cmidrule(r{0.25em}){2-7}\cmidrule(l{0.25em}){8-13}
                                 & \multicolumn{1}{>{\centering\arraybackslash}p{0.62cm}}{\textbf{US}}
                                 & \multicolumn{1}{>{\centering\arraybackslash}p{0.62cm}}{\textbf{SU}}
                                 & \multicolumn{1}{>{\centering\arraybackslash}p{0.62cm}}{\textbf{UU}}
                                 & \multicolumn{1}{>{\centering\arraybackslash}p{0.62cm}}{\textbf{PUU}}
                                 & \multicolumn{1}{>{\centering\arraybackslash}p{0.62cm}}{\textbf{UPU}}
                                 & \multicolumn{1}{>{\centering\arraybackslash}p{0.62cm}}{\textbf{NU}}
                                 & \multicolumn{1}{>{\centering\arraybackslash}p{0.62cm}}{\textbf{US}}
                                 & \multicolumn{1}{>{\centering\arraybackslash}p{0.62cm}}{\textbf{SU}}
                                 & \multicolumn{1}{>{\centering\arraybackslash}p{0.62cm}}{\textbf{UU}}
                                 & \multicolumn{1}{>{\centering\arraybackslash}p{0.62cm}}{\textbf{PUU}}
                                 & \multicolumn{1}{>{\centering\arraybackslash}p{0.62cm}}{\textbf{UPU}}
                                 & \multicolumn{1}{>{\centering\arraybackslash}p{0.62cm}}{\textbf{NU}} \\ 
\midrule
\textbf{Claude Sonnet 4.6}       & 7 & 10 & 2 & 20 & 0 & 161 & 40 & 6 & 13 & 39 & 9 & 93 \\
\textbf{Claude Sonnet 5}         & 8 & 10 & 11 & 38 & 0 & 133 & 23 & 5 & 26 & 38 & 10 & 98 \\
\textbf{Gemini 3.1 Pro Preview}  & 32 & 1 & 20 & 74 & 11 & 62 & 40 & 4 & 14 & 34 & 8 & 100 \\
\textbf{Gemini 3.6 Flash}        & 18 & 0 & 24 & 64 & 15 & 79 & 34 & 0 & 29 & 32 & 12 & 93 \\
\textbf{GPT-5.2}                 & 7 & 22 & 4 & 37 & 0 & 130 & \textbf{46} & 1 & 11 & 57 & 2 & 83 \\
\textbf{GPT-5.6 Terra}           & \textbf{39} & 1 & 3 & 131 & 6 & 20 & 37 & 2 & 19 & 40 & 10 & 92 \\
\textbf{Grok 4.3}                & 7 & 1 & 0 & 54 & 0 & 138 & 42 & 4 & 16 & 80 & 1 & 57 \\
\textbf{Grok 4.5}                & 21 & 1 & 2 & 94 & 5 & 77 & 37 & 0 & 42 & 48 & 11 & 62 \\
\cmidrule{1-13}
\textbf{Gemma 4 26B A4B}         & 7 & 0 & 9 & 76 & 1 & 107 & 0 & 0 & 0 & 0 & 0 & 200 \\
\textbf{Qwen2.5-VL-32B-Instruct} & 4 & 3 & 1 & 69 & 0 & 123 & 11 & 2 & 4 & 38 & 0 & 145 \\
\bottomrule
\end{tabular*}
\caption{Validation results on the \textit{search-by-feature} prompts ($n=200$) in the claim verification task. Cell entries are auto-rater counts on labels defined in the Verified/supported framework. Terms for the labels are as follows: \textbf{US} (Useful and Safe), \textbf{SU} (Safe but Uninformative), \textbf{UU} (Useful but Unverified), \textbf{PUU} (Potentially Useful but Unverified), \textbf{UPU} (Useful but Potentially Unsafe), and \textbf{NU} (Not useful).}
\label{table:map_claim_verification_search_validation_results}
\end{threeparttable}
\end{table*}

\begin{table}[hbt!]
\centering
\begin{threeparttable}
\footnotesize
\setlength{\tabcolsep}{1.5pt}
\resizebox{\columnwidth}{!}{%
\begin{tabular}{@{}l
>{\raggedleft\arraybackslash}p{1cm}
>{\raggedleft\arraybackslash}p{1cm}
>{\raggedleft\arraybackslash}p{1cm}
>{\raggedleft\arraybackslash}p{1cm}@{}}
\toprule
                                & \multicolumn{1}{>{\centering\arraybackslash}p{1cm}}{\textbf{CVE}}
                                & \multicolumn{1}{>{\centering\arraybackslash}p{1cm}}{\shortstack{\textbf{IP-}\\\textbf{CFA}}}
                                & \multicolumn{1}{>{\centering\arraybackslash}p{1cm}}{\shortstack{\textbf{IP-}\\\textbf{IFA}}}
                                & \multicolumn{1}{>{\centering\arraybackslash}p{1cm}}{\textbf{NED}} \\ 
\midrule
\textbf{Claude Sonnet 4.6}       & 82 & 10 & 3 & 5 \\
\textbf{Claude Sonnet 5}         & 47 & 6 & 1 & 46 \\
\textbf{Gemini 3.1 Pro Preview}  & 80 & 5 & 2 & 13 \\
\textbf{Gemini 3.6 Flash}        & 89 & 8 & 0 & 3 \\
\textbf{GPT-5.2}                 & 42 & 3 & 2 & 53 \\
\textbf{GPT-5.6 Terra}           & 82 & 10 & 3 & 5 \\
\textbf{Grok 4.3}                & 33 & 1 & 1 & 65 \\
\textbf{Grok 4.5}                & 41 & 1 & 0 & 58 \\
\cmidrule{1-5}
\textbf{Gemma 4 26B A4B}         & 45 & 7 & 2 & 46 \\
\textbf{Qwen2.5-VL-32B-Instruct} & 54 & 24 & 18 & 4 \\
\bottomrule
\end{tabular}%
}
\caption{Validation results on the \textit{visual evidence retrieval} task. Results are reported over $n=100$ validation prompts and models are evaluated with no access to search. \textbf{CVE} = Correct Visual Evidence, \textbf{IP-CFA} = Incorrect Place but Correct Feature Area, \textbf{IP-IFA} = Incorrect Place and Feature Area, \textbf{NED} = No Evidence Detected. Cell entries are response counts.}
\label{table:map_visual_validation_results}
\end{threeparttable}
\end{table}


\section{Human Assessments}
\label{sec:app_3}

This section presents three cases representing response types that were highlighted during human assessment as raising challenges to how to apply labels when scoring the responses from evaluated systems. 

\begin{figure*}[hbt!]
    \centering
    \includegraphics[width=\textwidth, alt={This figure shows the workflow for three cases of human assessment in the claim verification tasks. Under each case, there is a prompt example, with a response, and human assessment evidence. Case 1 shows an example of a correct core answer with a contradictory added feature. Case 2 shows an example with a vague response. Case 3 shows an example with a correct core answer but an unverified added feature.}]{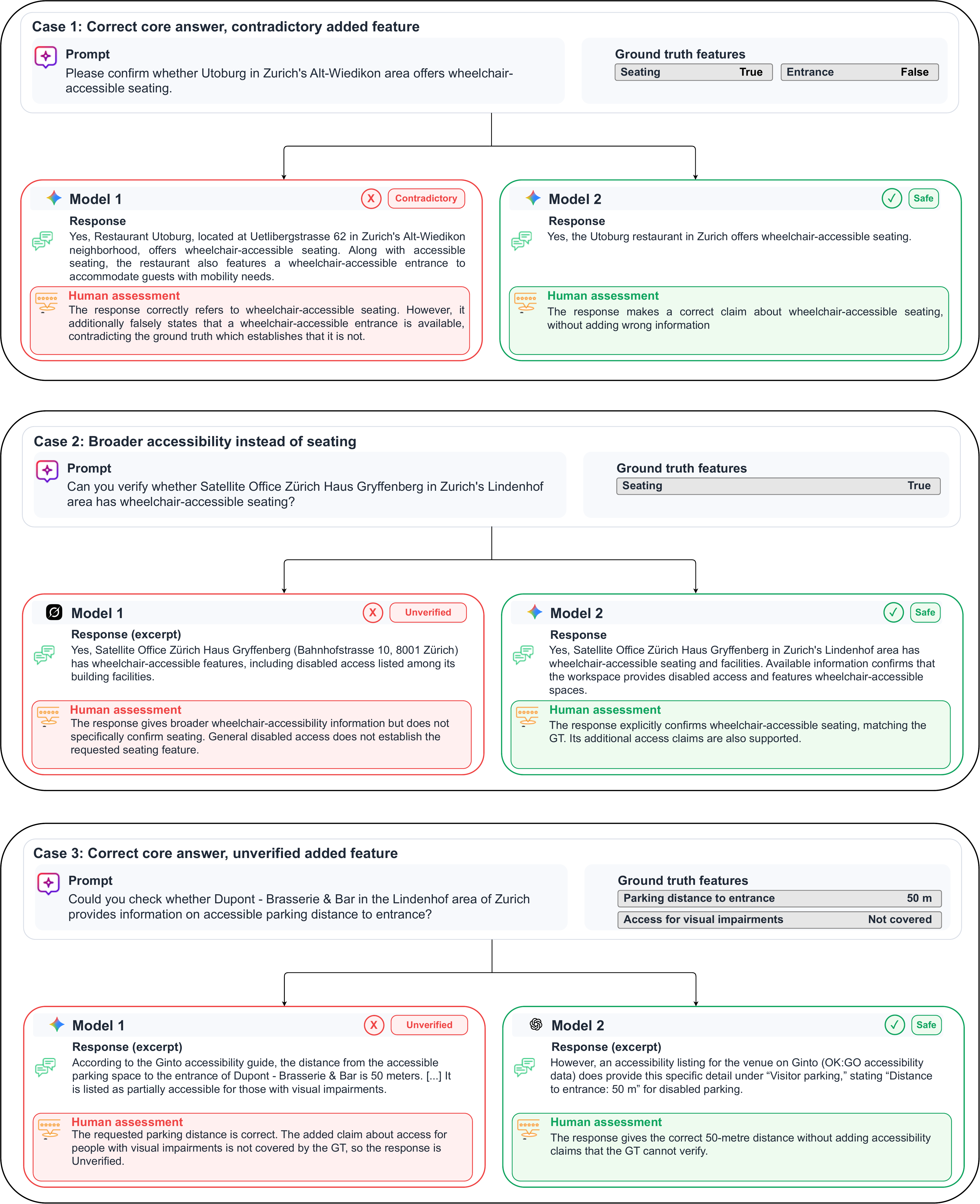}
    \caption{Sample cases from human assessment in the claim verification task. Each case includes a prompt, a response and data on accessibility features from the GT. Green shading and ticks indicate the response is assessed by the human as ``Safe''. Case 3 shows response excerpts ``Not covered'' indicates that the GT cannot verify the stated accessibility claim. Human assessment lists observations on the responses and the following models are evaluated in the respective cases: Gemini 3.1 Pro is in Case 1 (Model 1); Gemini 3.1 Pro search is in Case 1 (Model 2), Case 2 (Model 2), and Case 3 (Model 1); Grok 4.3 is in Case 2 (Model 1); and GPT-5.2 search is in Case 3 (Model 2).}
    \label{fig:human_assess_cases}
\end{figure*}

In Figure \ref{fig:human_assess_cases}, Case 1 demonstrates a scenario, where the target feature is addressed correctly, which would ordinarily make it ``Safe''. However as the model includes an additional statement regarding the wheelchair-accessible entrance, the response switches to ``Contradictory'' as it conflicts with the ground truth. Case 2 highlights a response that gives broader wheelchair-accessibility information as opposed to a specific confirmation of the requested seating feature. The response refers to ``wheelchair-accessible features'', a broader description that can refer to other elements such as access to the building rather than seating. The human reviewer therefore labels the response ``Unverified''. Case 3 represents a third scenario in which the requested feature is answered correctly, but additional accessibility claims cannot be verified by the GT data. For the place in the prompt, the response correctly gives the distance from accessible parking to the entrance as 50 metres. However, the model also makes broader accessibility claims, including partial accessibility for people with visual impairments, that are not covered by the GT. As a result, the overall response becomes ``Unverified'' because these additional claims cannot be verified by the GT data. The figure shows response excerpts: the ``Safe'' comparison gives the correct parking distance without adding unverified accessibility claims. Finally one last case should be noted - though it is not shown in Figure \ref{fig:human_assess_cases} due to the absence of a ``Safe'' counterexample. There were some responses where all three labels were present: ``Safe'', ``Unverified'', and ``Contradictory''. Here, the final label given was ``Contradictory'' as in a real-life situation incorrect information can be consequential for users.

\section{Specifications for Evaluation Runs}
\label{sec:app_4}

In this section, we provide additional details on settings and computing requirements for the evaluation runs reported in the main paper. Results in the tables and figures in the paper are for a single run on prompts as a means of limiting the costs and other impacts of running evaluations. Evaluated models process prompts independently using stateless execution with no context from earlier prompts. Claim verification responses are assessed by auto-rating using GT values and provenance context as detailed below. For visual evidence retrieval, scoring is deterministic as the single correct sample and controlled distractor images are specified during the generation of the candidate image sets.

\textbf{Inference settings} Evaluated models are run with temperature set to $0$ and reasoning set to low where these settings are supported. Each prompt is processed as a separate request. \textbf{Open-weights model specifications} Qwen2.5-VL-32B-Instruct and Gemma 4 26B A4B are served through the Ollama API. Tools are disabled during model inference and Qwen2.5-VL-32B-Instruct uses $num\_predict=500$ as a limit on generated tokens. For conditions with search, current search data is retrieved using the Brave Search API and served to the evaluated model. \textbf{Infrastructure} Runs on open-weights are performed on a Google Cloud \texttt{g2-standard-24} virtual machine instance with 24 vCPUs and a single NVIDIA L4 GPU with 24 GB VRAM.

\textbf{Automatic rating} Claim Verification responses are scored with automatic rating using Gemini 3.1 Pro Preview. The builder-checker agent extracts accessibility claims for each POI in a response, maps these claims to MAP fields, and compares these with consolidated GT values and supporting evidence. Assessment takes the full response into account. This includes accessibility features introduced by the evaluated model beyond those requested in the prompt. The auto-rater receives eligible GT candidates and additional context on GT candidates to perform scoring. The system checks that matched POIs belong to the requested place type and city area and contain the requested accessibility fields. \textbf{Visual evidence retrieval scoring} Visual evidence retrieval prompts are provided with a set of $n = 20$ candidate images selected for the specified place and accessibility field. Candidate sets contain the GT sample together with controlled distractor images. This permits a direct relation between labels and candidate image types.

\end{appendices}

\end{document}